\documentclass{INTERSPEECH2023}
\usepackage[table,xcdraw]{xcolor}
\usepackage{adjustbox}
\usepackage{caption}
\usepackage{graphicx}
\usepackage{hyperref}
\usepackage{dblfloatfix}
\usepackage{booktabs}
\hypersetup{
    colorlinks=true,
    linkcolor=blue,
    filecolor=magenta,      
    urlcolor=cyan,
    }

\interspeechcameraready

\title{A Parameter-Efficient Learning Approach to Arabic Dialect Identification with \\ Pre-Trained General-Purpose Speech Model}

\name{Srijith Radhakrishnan$^{1,2,4}$, Chao-Han Huck Yang$^{1,3}$, Sumeer Ahmad Khan$^{1,4}$\\
Narsis A. Kiani$^5$, David Gomez-Cabrero$^1$, Jesper N. Tegner$^{1,4,5}$}
\address{
  $^1$King Abdullah University of Science and Technology, Saudi Arabia\\
  $^2$Manipal Institute of Technology, India;
  $^3$Georgia Institute of Technology, USA;
  $^4$SDAIA-KAUST Center of Excellence in Data Science and Artificial Intelligence;
  $^5$Karolinska Institutet, Sweden}
  
\email{\{srijith.radhakrishnan,sumeer.khan,jesper.tegner\}@kaust.edu.sa, huckiyang@gatech.edu}

\begin{document}

\maketitle

\begin{abstract}

In this work, we explore Parameter-Efficient-Learning (PEL) techniques to repurpose a General-Purpose-Speech (GSM) model for Arabic dialect identification (ADI). Specifically, we investigate different setups to incorporate trainable features into a multi-layer encoder-decoder GSM formulation under frozen pre-trained settings. Our architecture includes residual adapter and model reprogramming (input-prompting). We design a token-level label mapping to condition the GSM for Arabic Dialect Identification (ADI). 
We achieve new state-of-the-art accuracy on the ADI-17 dataset by vanilla fine-tuning. We further reduce the training budgets with the PEL method, which performs within 1.86\% accuracy to fine-tuning using only 2.5\% of (extra) network trainable parameters. Our study demonstrates how to identify Arabic dialects using a small dataset and limited computation with open source code at \href{https://github.com/Srijith-rkr/KAUST-Whisper-Adapter}{\textit{https://github.com/Srijith-rkr/KAUST-Whisper-Adapter.}}

\end{abstract}

\section{Introduction}

Dialect identification~\cite{zissman1996automatic, ambikairajah2011language} (DI) amounts to identifying similar dialects belonging to the same language family. It is a specific case of language identification~\cite{watanabe2017language} (LID) task. However, DI is more challenging than LID owing to the fact that dialects share similar acoustic and linguistic characteristics compared to different languages. Very minute differences in pronunciation~\cite{5373245} and accent are used as cues to identify dialects. Moreover, DI does not share the advantage of publicly available speech recognition models pre-trained on large speech data corpora for network initialization. Despite these challenges, DI remains relatively unexplored compared to LID.

\begin{figure}
    \centering
    \includegraphics[width=0.46\textwidth]{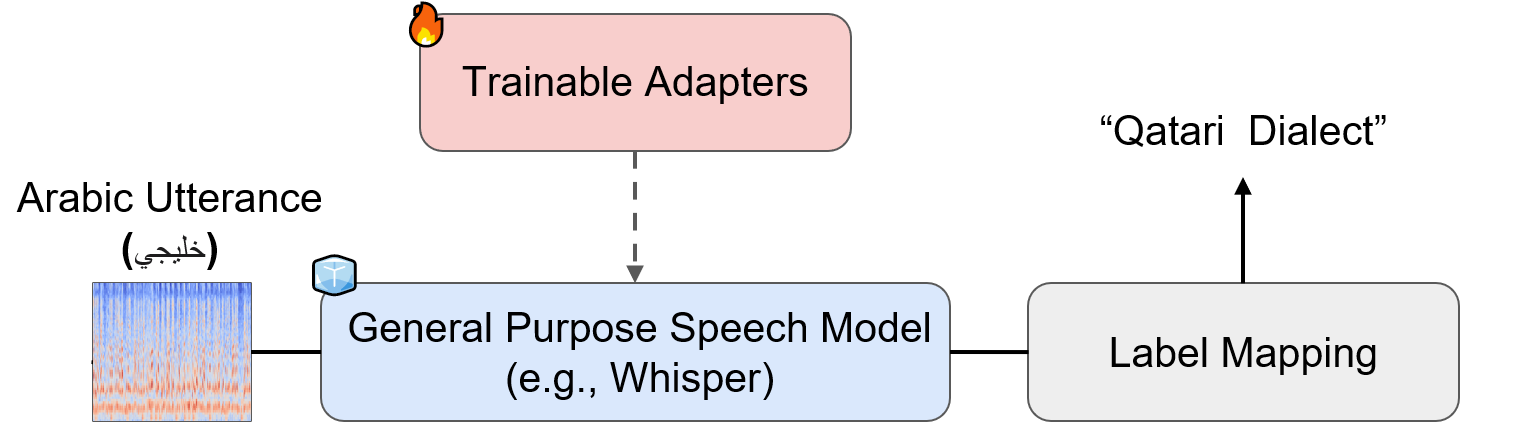}%
    \caption{Overview of proposed parameter-efficient learning framework for Arabic dialect identification 
 building upon parameter-efficient learning~\cite{hetowards} and label mapping~\cite{voice2series}.}
    \label{fig:overall_diagram}
\end{figure}

In this study, we leverage upon a recent open-access and general-purpose speech recognition architecture, Whisper~\cite{Whisper}, pre-trained on a large speech corpus from OpenAI, to address DI in resource-constrained and data-limited conditions. We use Parameter-Efficient Learning~\cite{yang2023english, hetowards, chen2023chapter, fu2022adapterbias} (PEL) to adapt a large pre-trained model by training small additive modules embedded into the frozen pre-trained model. By doing so, we require less training time and computing resources to fine-tune the model for DI. Figure \ref{fig:overall_diagram} is a schematic of the proposed parameter-efficient learning framework. 

We choose to perform DI in Arabic owing to its substantial regional variations and the widespread use of Arabic as an official language in over 22 countries~\cite{zaidan2014arabic}. Notably, significant differences exist between the standard written form, referred to as Modern Standard Arabic, and the local colloquial dialects spoken in each region. Interestingly, not all dialects are  mutually intelligible.\newline

In this study, we present several contributions: (1) Firstly, we introduce the novel use of Parameter-Efficient-Learning (PEL) for this task, marking the first application of this approach to Arabic dialect identification~\cite{zaidan2014arabic}.
(2) We investigate \textit{different designs} to incorporate trainable features into a multi-layer encoder-decoder frozen model. (3) We achieve new state-of-the-art accuracy on the \textbf{official} testing and development sets of ADI-17~\cite{ADI17} dataset using only 30.95\% of the training data. (4) Lastly, we demonstrate that our PEL method achieves equivalent performance to full fine-tuning using only \textbf{2.5}\% of (extra) network parameters.

\section{Related work}

\subsection{Existing works on Dialect prediction}
Several works exist in applying Natural Language Processing to the Arabic text due to sufficient open-source Arabic textual data from multiple sources, such as Newspaper articles \cite{s2}, and Twitter \cite{s22}. Unfortunately, minimal open-source Arabic speech data is available, as reported in \cite{review}. Most existing Arabic dialect identification methods use machine learning and deep learning methods. \cite{ali16_interspeech} used phonetic and lexical features obtained from a speech recognition system combined with a multi-class support vector machine~\cite{hearst1998support} to identify Arabic dialects. \cite{MGB3} used a Siamese neural network along with i-vector post-processing to learn similarities and dissimilarities among Arabic dialects.

\subsection{Parameter-Efficient Learning}
Large pre-trained models have been very successful at various tasks in natural language processing. Parameter-efficient learning is a research direction that aims to reduce the computational cost by adapting large pre-trained models to downstream tasks by updating only a subset of (extra) parameters. In this section, we introduce several state-of-the-art parameter-efficient learning methods. 

\textbf{Residual Adapters} : Residual Adapters are small trainable blocks inserted between the layers of a transformer architecture \cite{houlsby,NIPS2017_e7b24b11,tomanek-etal-2021-residual, chang2022exploration}. They down-project the latent dimension from the previous layer and apply a nonlinear activation function, followed by an up-projection. A residual connection surrounds the adapter layer. This setup encourages parameter sharing between the frozen components and localizes all the weight updates to the adapter modules. 

\textbf{Neural Reprogramming}: Neural reprogramming can be used to repurpose a frozen pre-trained model to out-of-domain prediction tasks by adding trainable parameters to the input of the pre-trained model~\cite{voice2series,yen2021neural}. The frozen model is followed by a label-mapping strategy to map the source labels to the out-of-domain target labels. The trainable input noise aligns the latent distribution of the target domain with being more similar to that of the source domain, using the pre-existing decision boundaries of the frozen model. Neural reprogramming works well when the input size of the target data is comparably smaller than the source data, as demonstrated in \cite{voice2series,NR1, yang2023english}. 

\textbf{BitFit}: BitFit refers to BIas-Term FIne-Tuning, which only updates the bias-terms and the task-specific classification layer of the frozen pre-trained model \cite{bitfit}.

\textbf{Others}: Other state-of-the-art parameter-efficient learning techniques include LoRa \cite{lora},  in which trainable low-rank matrices are embedded inside the transformer attention layers to approximate parameter updates and speech prompting methods ~\cite{chang2022exploration} as one approach shared similar motivation to input reprogramming, such trainable input and label mapping.

\section{Method}
We evaluate two fine-tuning strategies. First vanilla fine-tuning, in which we fine-tune different components of the network parameters. Next, parameter-efficient fine-tuning, in which we train (extra) modules added to the network architecture and investigate their performance. 

\begin{figure}
    \centering
    \includegraphics[width=0.49\textwidth]{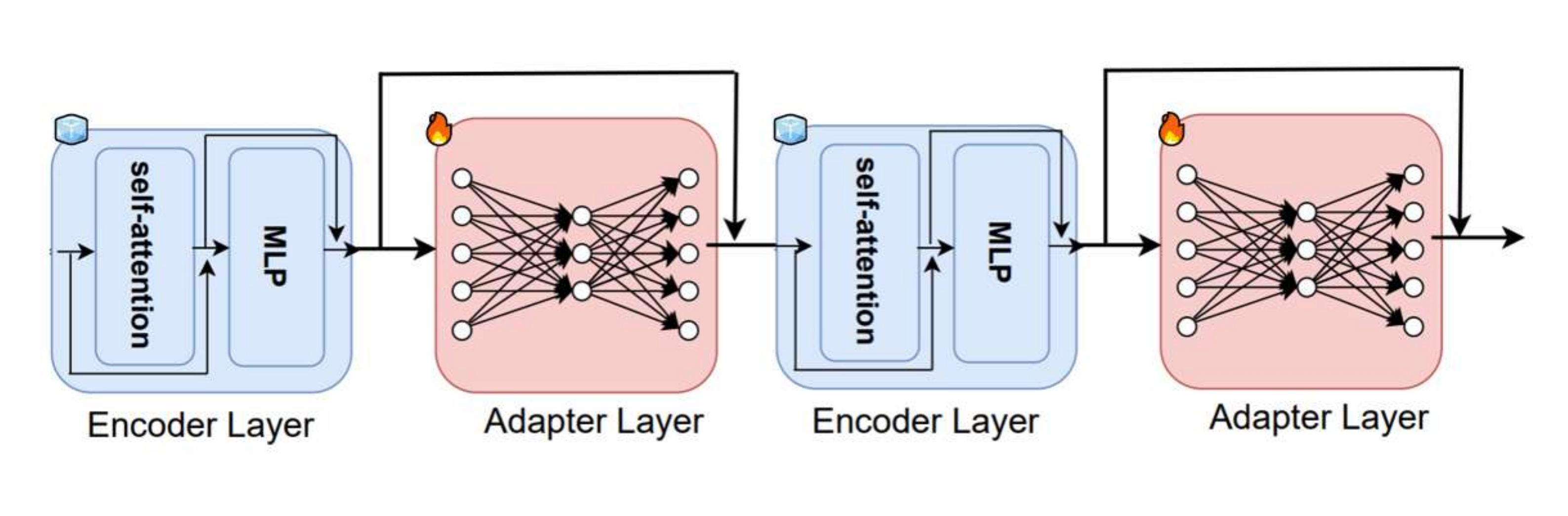}%
    \captionsetup{skip=11pt}
    \caption{Illustration of the transformer architecture embedded with adapter layers.}%
    \vspace{-4mm}
    \label{fig:adapter}
\end{figure} 

\subsection{Vanilla Fine-tuning}
To incorporate the new dialect classes into the network architecture, we append the new classes to the pre-existent multilingual tokenizer and modify the token embedding matrix $W_{e} \in \mathbb{R}^{t\times n}$   to   $W^{^{\prime}}_{e} \in \mathbb{R}^{(t+d)\times n}$. Here $t$ and $n$ are the dimensions of the tokenizer and network, respectively, and $d$ denotes the number of dialects. The weights of 
$W_{e}$ are then copied with random padding to $W^{^{\prime}}_{e}$ for initialization. A cross-entropy loss is used to fine-tune different components of the network.

\subsection{Input Reprogramming}
We begin our experiments in parameter-efficient fine-tuning with Input reprogramming. If $x$ is the input to the pre-trained frozen model $W_{\Theta}\left ( . \right )$ parameterized by theta, the frozen mode predicts the dialect $\hat{y}$ as $W_{\Theta}\left ( x \right )  \rightarrow \hat{y}$. Input reprogramming aims to add trainable noise $x_{t}$ to the input $x$. In our application, the input to the frozen pre-trained model is the Log-Mel Spectrogram computed from 30 seconds of Arabic speech. We add trainable parameters of the same dimensions as $x$ to minimize the prediction error between $\hat{y}$ and the true label while not updating ${\Theta}$, the parameters of the model as $W_{\Theta}\left ( x  + x_{t}\right )  \rightarrow \hat{y}$

\subsection{Latent Space Efficient Learning}
We insert small trainable modules called adapters between the encoder layers of our model as shown in Fig \ref{fig:adapter}. These adapter layers contain a linear down projection of the latent input dimensions $n$ to a bottleneck dimension $b$ using $W_{dp} \in \mathbb{R}^{n\times b} $. We apply $g\left ( . \right )$ , the GELU \cite{gelu} activation function on the bottleneck dimension followed by an up projection with $W_{up} \in \mathbb{R}^{b\times n}$. Residual connections are applied around the adapter layer as represented in Equation \ref{eq1}. We compare the performance of multiple bottleneck dimensions $b$, set to $\frac{n}{2},\frac{n}{4}$, and $\frac{n}{8}$ against multiple sample sizes in our experiments.
\begin{equation}\label{eq1}
x \leftarrow x + g\left ( x .W_{dp} \right )W_{up} 
\end{equation}

\subsection{Token Mapping in Whisper}
We utilize many-to-one hard-label mappings to map the source label of the model to our target dialect classes, motivated by~\cite{voice2series}. Specifically, we randomly assign unique language tokens of the model to each dialect, sum over their logits, and apply the softmax function to calculate respective dialect probabilities. Performance significantly depreciated when we attempted to implement a trainable label mapping setup. 

\section{Experimental Setup}
In this section, we describe our experimental setup and present results for Arabic dialect identification on the ADI-17 dataset.
\subsection{General Purpose Speech Model}
We use Whisper$_\text{Base}$ \cite{Whisper} as the underlying pretrained model composed of an encoder-decoder architecture for our experiments. A multi-task general purpose speech model trained on 680,000 hours of multilingual data from 99 languages ensuring rich representational knowledge. The input to the model is the log magnitude Mel spectrogram representation computed from 30 seconds of an audio clip. We pad the audio clip with zeros if the duration is less than 30 seconds. More details can be found at \cite{Whisper}. For training we use Adam optimizer for all tasks. We explore learning rates of  {{$1\mathrm{e}{-2}$, $1\mathrm{e}{-3}$, $1\mathrm{e}{-4}$}}, and utilize the optimal learning rate with a linear learning rate scheduler to train for 50 epochs until convergence. We trained our models using 4 V100 GPUs with an adequate batch size of 64 and 0.1 weight decay. Our implementation and pre-trained weights are open source at \href{https://github.com/Srijith-rkr/KAUST-Whisper-Adapter}{\textit{https://github.com/Srijith-rkr/KAUST-Whisper-Adapter}}.

\begin{table*}[t]
\caption{An overview of the model performance of fine-tuning and parameter-efficient learning on utterances over $20$ seconds as in standard ADI-17 setup. The development and test accuracy were evaluated in the official setup and outperformed three previous state-of-the-art results~\cite{ADI17, clusterinADI, transformerADI17} using its official dev and test settings. Noted that directly using frozen pre-trained Whisper (second row) with label mapping performs 2.52\% to 4.44\% in the dialect identification task, which motivates its demand of efficient model designs.
}
\label{parameter}
\centering
\begin{tabular}{cccccc}
\hline
{\textbf{Method}} & \textbf{Trainable Para. ($\downarrow$)} & \textbf{Trainable Ratio (\%)} & \textbf{Dev. Acc. ($\uparrow$)} & \textbf{Test Acc. ($\uparrow$)} & \textbf{Utility Score($\uparrow$)}\\ \hline
Frozen Whisper$_\text{Base}$~\cite{Whisper}           & 0  & 0\%   & \cellcolor[HTML]{FFCCC9}2.52\% & \cellcolor[HTML]{FFCCC9}4.44\% & - \\ \hline
Full fine-tuning          & 71.8M  & 100\%   & 95.55\% & 93.34\% & 11.88 \\ 
Encoder fine-tuning       & 18.9M  & 26.32\% &  \textbf{96.39}\%  &  \textbf{95.01}\%  & 13.06   \\ 
Decoder fine-tuning       & 52M    & 72.42\% & 95.07\% &   93.75\%   & 12.15\\ \hline
BitFit                    & 75.8K  & 0.10\%  &  59.01\% &   57.68\%  & 11.82  \\ 
Encoder BitFit            & 32.3K  & 0.04\%  & 44.59\% &  41.83\%     & 9.28  \\ 
Decoder BitFit            & 43.5K  & 0.06\%  &  36.78\%  &   39.42\%  & 8.50\\ \hline
Input Reprogramming & 240K   & 0.33\%  & 27.04\% & 27.91\% & 5.19 \\ 
Adapters-64              & 642K   &  0.89\%  & 92.79\% &   89.47\% & \cellcolor[HTML]{9AFF99}\textbf{15.41}  \\ 
Adapters-128              & 1M   &   1.39\% &    93.99\%   &   91.50\%  & 15.25    \\ 
Adapters-256              & 1.8M   & 2.50\%  & \cellcolor[HTML]{9AFF99}\textbf{95.55}\% &  \cellcolor[HTML]{9AFF99}\textbf{93.15}\%  & 14.89   \\ \hline \hline

ADI-17 \cite{ADI17}              & 13.1M  & - & 93.70\% &  90.4\%  & 13.17 \\ 
CNN+Transformer \cite{transformerADI17}               & 64.3M  & -  & 94.04\% &  93.06 & 11.92   \\ 
Supervised clustering \cite{clusterinADI}             & $>$50M  & -  & 95.09\% &  94.35\%   & $\sim$12.25  \\  \hline

\end{tabular}
\vspace{-6mm}
\label{tab:1}
\end{table*}

\subsection{Dataset}
We evaluate our findings on the ADI-17 dataset \cite{ADI17} released as part of the MGB5 challenge. The dataset contains audio data for 17 Arabic dialects collected from YouTube. The training set contains 3033 hours of audio data. The dev and test set contain 58 hours of audio data. The training set of ADI17 is unbalanced in terms of the number of utterances and hours of data. For example, the Iraq (IRQ) dialect has 815.8 hours of training data, whereas the Jordan (JOR) dialect contains only 25.9 hours of training data. The audio clips have varying lengths, between 1 second to 26 minutes. To deal with class imbalance during training, we over-sample random windows of 30 seconds of audio data from clips from the minority classes containing more than 20 seconds in duration. The dataset is further divided into three subcategories in terms of duration, Short duration ($<$ 5s), medium duration (5-20s), and long duration ($>$ 20s). We perform our experiments using the long duration ($>$ 20s) subsection of the dataset since we only use a fraction of the dataset to fine-tune different transfer learning methods to evaluate the effect of data on performance. The sample sizes and their duration are found in Table \ref{sample}.

\begin{table}[]
\caption{Dataset sample statistics}
\label{sample}
\begin{tabular}{ccc}
\hline
\textbf{\#samples per class} & \textbf{\#hours per class} & \textbf{fraction of ADI-17} \\ \hline
0.5k                & 4.1                        & 2.33\%                       \\
1k               & 8.2                        & 4.60\%                       \\
2k               & 15.5                       & 8.69\%                       \\
5k               & 32.8                       & 18.40\%                      \\
10k              & 53.1                       & 30.95\%                      \\ \hline

\end{tabular}%
\end{table}
\subsection{Baseline}
 The baseline reported by ADI-17 \cite{ADI17} and the performance of other methods on ADI-17 with utterances over 20 seconds have been reported in Table \ref{tab:1}. ADI-17 \cite{ADI17} utilized a convolutional neural network-based (CNN) architecture with a softmax output. The \cite{transformerADI17} used a fusion of transformer-based architecture and CNN with downsampling. \cite{clusterinADI} used a supervised clustering-based algorithm with triplet loss. The results in the table show that all three methods perform well in terms of accuracy. However, there are notable differences in the number of parameters used by each method, with ADI-17 \cite{ADI17} utilizing the fewest parameters at 13.1M and Supervised clustering \cite{clusterinADI} requiring over 50M parameters.

\subsection{Performance Studies}
Whisper$_\text{Base}$ is an encoder-decoder-based transformer~\cite{vaswani2017attention} architecture with a total trainable parameter of $70$M. We fine-tuned the encoder and decoder separately to examine the impact of the architecture on DI. Similarly, we evaluate the same setup for bias-only fine-tuning. 
The performance of fine-tuning with $10,000$ samples and the number of fine-tuned parameters is reported in Table \ref{tab:1}. Fine-tuning only the encoder achieves new state-of-the-art accuracy of 95.01\% on ADI-17 while using only 30.95\% of the training set.
Fine-tuning the decoder alone yields comparable results to fine-tuning the entire model, but fine-tuning only the encoder achieves better results.
However, fine-tuning the encoder bias terms performs significantly better than fine-tuning decoder bias terms. Input reprogramming does not perform well compared to bias-only fine-tuning, although they share a similar number of parameters.\newline

\begin{figure*}[htb!]
    \centering
    \includegraphics[trim=0 0 0 30mm,clip, width=0.98\textwidth]{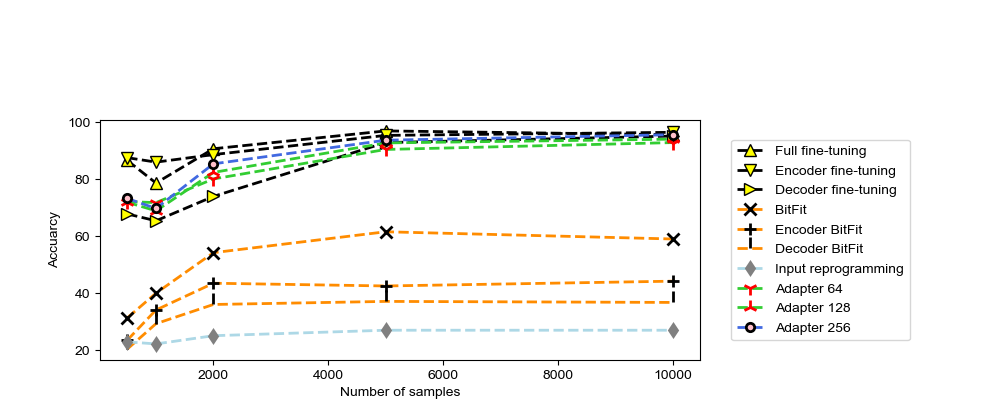}
    \caption{Performance comparison of fine-tuning methods using the development set against the number of samples per class.}
    \label{fig:dev}
\end{figure*} 

\begin{figure*}[htb!]
    \centering
    \includegraphics[trim=0 20mm 0 20mm,clip,width=0.99\textwidth]{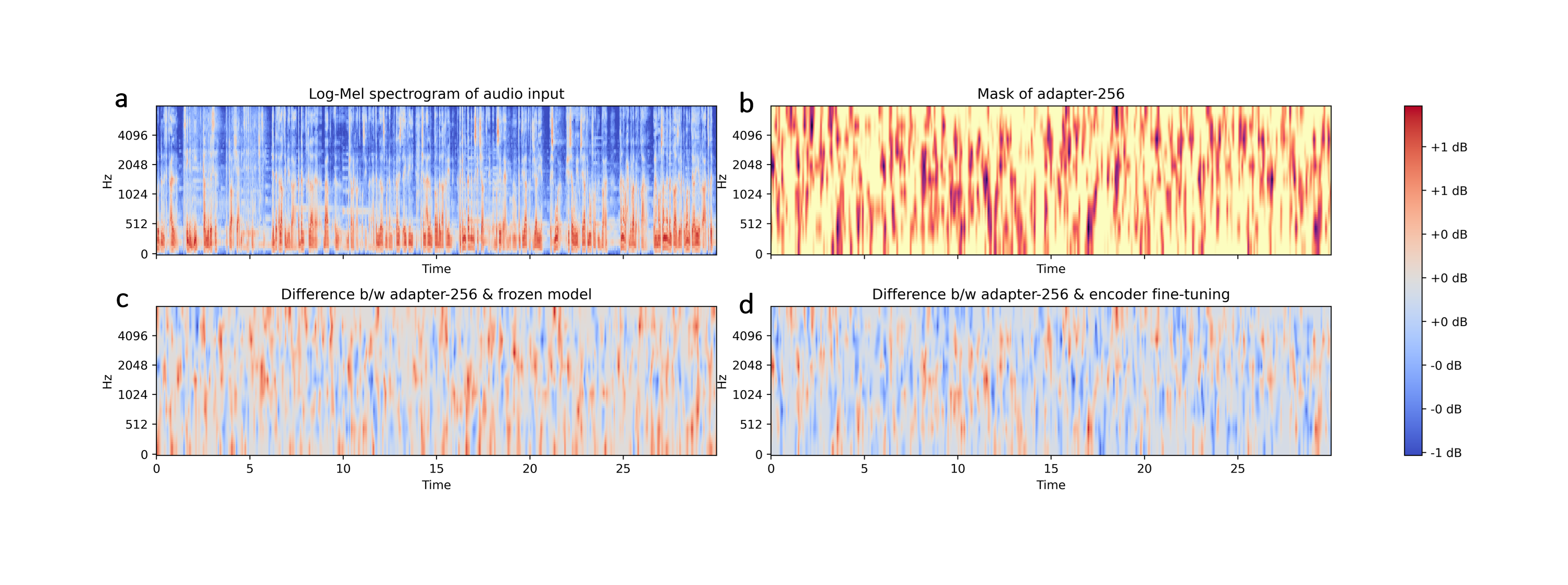}
\caption{(a) Log-Mel spectrogram of the audio input, (b) Mask of the adapter-256 model, (c) Difference between the masks of adapter-256 and frozen model and (d) Difference between the masks of adapter-256 and fine-tuned encoder model.}
    \label{fig:spec}
\end{figure*}

Whisper$_\text{Base}$ has $512$ latent dimensions from its transformer encoder. Thus, we train our residual adapter layers with bottleneck dimensions of $64$, $128$, and $256$ to examine the impact of the bottleneck dimensions on performance.\\ We also tested (1) adapters with bridge connections, (2) adapters with dense connections \cite{densenet}, (3) adapters without residual connections, (4) adapters with self-attention backbone, (4a) with Gelu activation, (4b) with residual connections. We observed that the setup in Section 3.3 had the best performance. The adapter approach with $256$ latent dimensions has the same performance as fully fine-tuning the model, despite it updating only 2.5\% of the network parameters. This is advantageous since a separate copy of all model parameters does not have to be created for each downstream task of Whisper. Instead, only the adapter weights need to be stored for each downstream task.  \newline 

The performance comparison of fine-tuning methods on the development set against the number of samples per class is displayed in Figure \ref{fig:dev}. Here we observe that although encoder fine-tuning, decoder fine-tuning, and full fine-tuning achieve similar performance against $10,000$ samples per class, encoder fine-tuning performs significantly better with a lower number of samples. We conjecture that this is because only the encoder of Whisper$_\text{Base}$ processes the audio input. The decoder receives the output of the encoder. A similar trend is observed with bias-only fine-tuning.
Furthermore, the adapter methods with $64$, $128$, and $256$ latent dimensions display comparable performance, with minor improvements in performance as the bottle-neck dimensions increase. Notably, the adapter methods outperform decoder fine-tuning with fewer samples.\newline %

\subsection{Utility Discussion and Efficient Module Selection}
 In addition to accuracy, we also report Utility scores as defined in Equation (\ref{eq2}) to compare the efficiency against performance. The results of this analysis are presented in Table \ref{tab:1}.

\begin{equation}\label{eq2}
Utility\ score \left ( i \right )= \frac{Acc\_test\left ( i \right )}{log\left (N\text{of trainable parameters}\ \right )}
\end{equation}

The results in Table \ref{tab:1} show that adapters demonstrate the highest \textit{Utility} scores among all methods, as they perform well with a smaller number of trainable parameters. Specifically, adapter-$64$ (ninth row) achieves the optimal performance versus efficiency tradeoff, indicating that this method may be the most efficient approach for the task.

\subsection{Neural Saliency Analysis on Acoustic Features}
As a preliminary study investigating PEL learning experiments for Arabic dialect identification tasks, we aim to provide attribute-based interpretation based on frozen pre-trained Whisper model. Neural saliency methods \cite{perturbations,gradcam} provide interpretable intuitions behind black-box networks by analyzing the weight distribution over hidden neurons.
In this work, we employ a perturbation-based saliency map~\cite{perturbations} for feature-level interpretations.  This algorithm masks parts of the input to determine regions responsible for the classifier decision to understand better the patterns the model uses to predict Arabic dialects. The results of our analysis are presented in Fig \ref{fig:spec}.
Fig 4a represents the Log-Mel spectrogram of 30 seconds of Qatari dialectal speech. The mask of adapter-256 in Fig 4b emphasizes the parts of the input the model focuses on to make predictions. These regions are highlighted in red color. The difference between the masks of the adapter, frozen model, and fine-tuned encoder model have been plotted in Fig 4c and 4d. The bluish plot in Fig 4d compared to Fig 4c indicates that the adapter-256 and fine-tuned encoder model observe similar input regions compared to the adapter-256 and frozen Whisper.
\section{Conclusion}
This paper presents parameter-efficient approach for Arabic dialect identification, as one very first study on this low-resource application. The proposed method utilizes frozen pre-trained speech recognition models by incorporating trainable parameters at the input and latent space levels. By using only 30.95\% of the ADI-17 training data and 2.5\% of the model parameters, the proposed method achieves state-of-the-art accuracy on the ADI-17 benchmark, comparable to the performance of full fine-tuning. These results showcase the effectiveness of our efficient learning approach. 
Further research on this approach may enable dialect prediction and other tasks for low-resource and long-tail data~\cite{chtlow,labrador2023exploring}, which deserves more studies.
\section{Acknowledgements}
We gratefully acknowledge SDAIA-KAUST for supporting this work.
\clearpage
\bibliographystyle{IEEEtran}
\bibliography{mybib}

\end{document}